\documentclass[conference]{IEEEtran}
\usepackage{graphicx}
\usepackage{cite}

\usepackage{grffile}
\usepackage{cite}
\usepackage{booktabs}
\usepackage{amsmath,amssymb,amsfonts}
\usepackage{graphicx}
\usepackage{textcomp}
\usepackage[table,xcdraw]{xcolor}
\usepackage{xcolor}
\usepackage[caption=false]{subfig}
\usepackage{multirow}
\usepackage{algorithm}
\usepackage{algpseudocode}
\ifCLASSINFOpdf
\else
\fi
\begin{document}
%
\title{Bangla-Wave: Improving Bangla Automatic Speech Recognition Utilizing N-gram Language Models}

\author{Participants: \IEEEauthorblockN{Mohammed Rakib\IEEEauthorrefmark{1},
Md. Ismail Hossain\IEEEauthorrefmark{1}\\
Supervisors: Nabeel Mohammed\IEEEauthorrefmark{1}, Fuad Rahman\IEEEauthorrefmark{2}}
\IEEEauthorblockA{\IEEEauthorrefmark{1}Apurba-NSU R\&D Lab, Department of Electrical and Computer Engineering\\
North South University, Dhaka, Bangladesh\\}
\IEEEauthorblockA{\IEEEauthorrefmark{2}Apurba Technologies\\
440 N. Wolfe Rd., Sunnyvale, CA 94085, USA\\}
\IEEEauthorblockA{Email: \IEEEauthorrefmark{1}\{mohammed.rakib, ismail.hossain2018, nabeel.mohammed\}@northsouth.edu}
\IEEEauthorrefmark{2}\{fuad\}@apurbatech.com}


%


\maketitle



%
\IEEEpeerreviewmaketitle

\begin{abstract}
Although over 300M around the world speak Bangla, scant work has been done in improving Bangla voice-to-text transcription due to Bangla being a low-resource language. However, with the introduction of the Bengali Common Voice 9.0 speech dataset, Automatic Speech Recognition (ASR) models can now be significantly improved. With 399hrs of speech recordings, Bengali Common Voice is the largest and most diversified open-source Bengali speech corpus in the world. In this paper, we outperform the SOTA pretrained Bengali ASR models by finetuning a pretrained wav2vec2 model on the common voice dataset. We also demonstrate how to significantly improve the performance of an ASR model by adding an n-gram language model as a post-processor. Finally, we do some experiments and hyperparameter tuning to generate a robust Bangla ASR model that is better than the existing ASR models.  
\end{abstract}

\section{Introduction}
Bengali is one of the world's most diversified languages with about 300 million native speakers worldwide. However, documenting the Bengali language in any field is difficult even though it is a widely spoken language. The key to effectively documenting any task in this language lies in the capacity to translate spoken language. Therefore, the technically proficient and swift option is ``speech to text" or creating an ASR model. ASR modeling in Bangla has widespread implications. It can be used to generate closed captions in videos which aids a person with hearing loss to comprehend what is being said. Instead of typing, those who have problems using their hands can interact with computers using voice instructions leveraging ASR models. Additionally, Bengali ASR can be applied in a variety of fields where individuals use their hands to write this complex language for documents, including the legal system, business, education, and many others.

The acquisition of a sizable, readable, and diverse dataset is one of the main constraints in ASR model training. Given the unusual morphology and wide range of accents of the Bengali language, a substantial corpus with proper annotations is even more essential for large-scale deep learning models. A comprehensive dataset for the Bengali language is the Bengali Common Voice Speech Corpus \cite{commonvoice}. A total of 399 hours of transcribed audio recordings of spoken Bengali sentences from Bangladesh and India were acquired through community outreach and collaborative initiatives.

\begin{figure}[!t]
	\centering
    {\includegraphics[width=\linewidth]{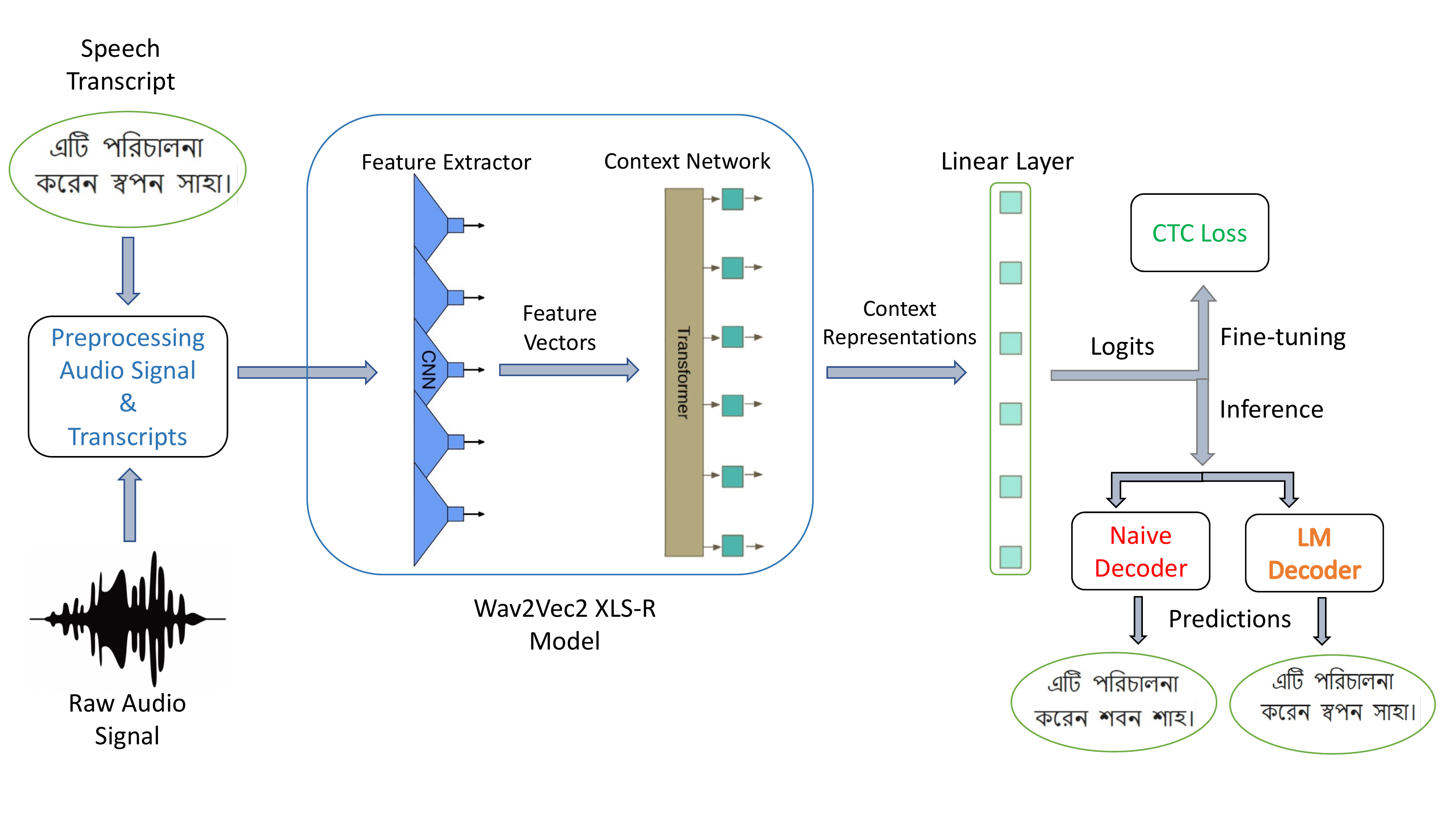}}  
	\caption{Here we can visualize the entire pipeline of our project. At first, we preprocess the raw audio signals and speech transcripts. Next, a feature extractor processes the signals to the Wav2Vec2 XLS-R model's input format. The model then takes these feature vectors as input and outputs a sequence of contextual representations. These sequences are then fed as input to the linear classification layer. While finetuning, the randomly initialized linear layer along with the transformer layers get trained using Connectionist Temporal Classification (CTC) loss. On the other hand, during inference, the linear layer outputs logits which are then fed to a decoder. For naive CTC decoding, there can be spelling mistakes since the ASR model only takes into account the context of the acoustic input. However, when an n-gram language model is combined, the performance improves significantly. This is shown by an example where we see that the predictions from the language model exactly match the ground truth speech transcript. On the other hand, the prediction directly from the naive CTC decoder is slightly off as it predicted some different characters.}
	\label{overview_diagram}
\end{figure}

Creating high-quality voice recognition applications is exceedingly challenging due to Bengali's wide variety of dialects and diverse speech sources. Creating a corpus with every dialect across all feasible areas is not practical. By learning from unlabeled training data, wav2vec 2.0 \cite{wav2vec2} pushes the boundaries and enables speech recognition systems to be employed in a wide range of domains and dialects. Similar to the Bidirectional Encoder Representations from Transformers (BERT), Wave2vec 2.0 is taught by anticipating speech units for masked audio segments. Speech audio differs greatly from other audio in that it captures a variety of recording elements without explicitly classifying them as words or other units. This issue is addressed by Wav2vec 2.0, which trains high-level contextualized representations using basic units that are 25ms long. Then these units are used to characterize various spoken audio recordings. As a result, algorithms for voice recognition can be developed that are more efficient than the best semi-supervised methods.

Facebook AI released a multilingual version of Wav2Vec2 called XLS-R \cite{xls-r}. A model's ability to learn speech representations that apply to several different languages is referred to as cross-lingual speech representations, or XLS-R. In this paper, we describe the implementation of ASR on the "Bengali Common Voice Speech" dataset utilizing the Wav2Vec2 XLS-R model. Fig-\ref{overview_diagram} gives an overview of our project. We carried out various types of preprocessing, finetuning, and post-processing and ultimately outperformed the current SOTA ASR models for Bangla. Concretely, our contributions are:

\textbf{Contributions:}
\begin{itemize}
	\item We have finetuned a pretrained ASR model on the Common Voice Bengali dataset and have achieved significant performance gains in transcribing audio signals. 
    \item We have added an n-gram language model as a post-processor to improve the accuracy of the transcriptions. 
\end{itemize}

\begin{figure*}[!ht]
	\centering
    {\includegraphics[width=\linewidth]{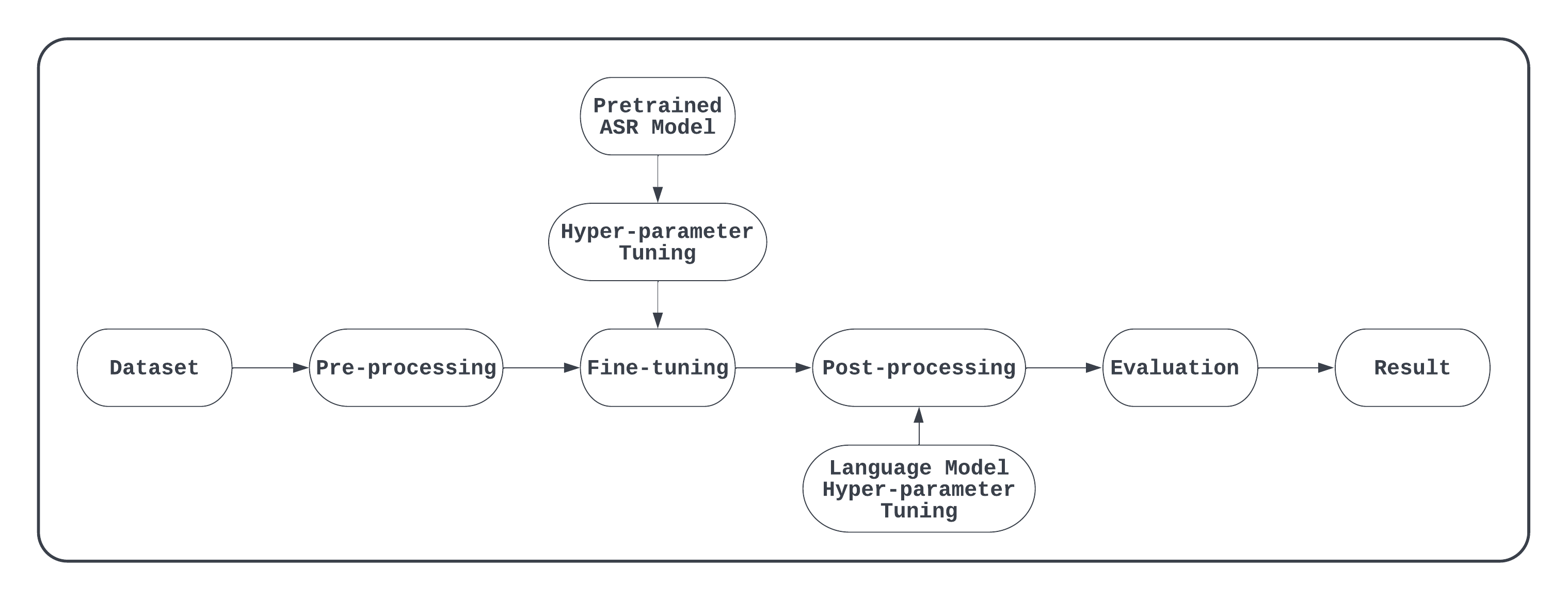}}  
	\caption{System Diagram of Bengali ASR model.}
	\label{system_diagram}
\end{figure*}

\section{Methodology}
Fig-\ref{system_diagram} depicts the system diagram of our project. At first, we preprocess the Bengali common voice dataset. Next, we selected the best available pretrained ASR model and performed hyperparameter tuning on a small subset of the common voice train set. After that, the pretrained model is finetuned on the preprocessed dataset. Subsequently, an n-gram language model \cite{ngram} is added as a post-processor after performing hyperparameter tuning on the parameters of the language model. Lastly, we perform an evaluation on the validation set and report the results.

\subsection{Dataset}
The Bengali Common Voice Speech Dataset \cite{commonvoice} is the largest and most diversified publicly available speech corpus for Bengali. It consists of 399hrs of speech recordings of which 57 hrs have been manually validated by one or more users. There are a total of 231,120 audio samples contributed by about 20k contributors of which 206950 samples are in the train set, 7747 samples in the validation set, 7747 samples in the test set, and the remaining dropped due to poor quality. The corpus has 135,625 unique sentences with 7.12 words on average for each sentence. The training data consists of about 84\% male samples and 16\% female samples. According to the authors of the dataset, it has more speaker, phoneme, and environmental diversity compared to Open-SLR\_slr53 which is the largest open-source dataset after Bengali Common Voice. 

\subsection{Preprocessing}
Speeches recorded as raw audio are less suitable for immediate use. The signals contain a lot of duplicate information, combined with noise that can seriously hinder recognition tasks. So we have to process the audio signal before feeding them as inputs to a model. For this, we use a feature extractor that will convert the speech signal to the model's input format and give a feature vector. Since ASR models transcribe speech to text, we also use a tokenizer that will convert the model's output format to text by decoding the predicted output classes to the output transcription. Since we will be using Connectionist Temporal Classification (CTC) loss to finetune our model, we categorized speech fragments into letters. We take all the unique letters from the training data and use them to create our vocabulary. To convert the string into a collection of characters, a mapping function is created that concatenates all transcriptions into a single lengthy transcription to give the desired output. Concretely, the preprocessing steps are as follows:

\begin{enumerate}
  \item \textbf{Transcript Normalization}: We normalized punctuation and characters that have multiple Unicode representations and also removed non-Bangla language characters using a Bangla Unicode normalizer \cite{bnunicode}.
  \item \textbf{Filtration}: The Bengali Common Voice dataset has some metadata which includes "upvotes" and "downvotes".  So, we removed samples with more downvotes than upvotes. We also removed samples if they had an audio duration greater than 20secs or less than 1sec. 
  \item \textbf{Define Token Class}: We ensured that all the missing Bengali vocabs and necessary punctuations from the Common Voice dataset were added as a token class.
  \item \textbf{Feature Extraction}: We have applied z-score normalization (making mean zero and standard deviation one) to the audio signals. Besides, we have resampled audio signals to 16kHz and lastly converted them to a one-dimensional array. In this way, the raw audio signals are processed to the model's input format by our feature extractor.
\end{enumerate}

\subsection{Pretrained Model}
To select the best pretrained model for finetuning, we looked up various state-of-the-art models and selected the best one. Evaluating various pretrained models, we saw that the Wav2Vec2 XLS-R model by arijit \cite{arijit_2020} had the best performance on the validation set of the Bengali Common Voice dataset. The 300M parameter model was first pretrained by Facebook AI on 436k hours of unlabeled speech in 128 languages. The speech datasets include VoxPopuli \cite{voxpopuli}, MLS CommonVoice \cite{mls}, BABEL \cite{babel}, and  \cite{voxlingua}. This pretrained model was then trained on 196k utterances of labeled Bengali speech data by OpenSLR\_slr53. We have selected this model to train on Bengali Common Voice 9.0. 

\subsection{Fine Tuning}
Before beginning finetuning, we first did hyperparameter tuning on a small subset of the train set to find the optimal parameter values for training the Wav2Vec2 XLS-R model. We found the optimal learning rate to be $3e_-4$. The wav2vec2 XLS-R model contains a stack of CNN layers at the beginning which is used to extract contextually independent yet acoustically significant characteristics from the raw speech signal. According to the research by the authors of \cite{xls-r}, this portion of the model has already received enough training during pretraining and no longer requires fine-tuning. As a result, we have frozen these layers and have not calculated or updated the gradients for these parameter values. Finally, we started finetuning and trained the model for 15 epochs on the 200k training samples of the common voice dataset and selected the model with the lowest loss on the validation set. 

\subsection{Post-Processing \& Language model}
To improve the performance of our finetuned wav2vec2 model, we added an N-gram language model as a post-processor. An N-gram language model is a probabilistic model developed using a large body of text that can predict the most probable next word given a sequence of N-1 words \cite{ngram}. So, a 3-gram/trigram language model predicts the next word in the sequence by taking into context the previous two words. ASR models like wav2vec2 output predictions are solely based on the acoustic input which may contain noise. As a result, the output often contains spelling mistakes as it cannot differentiate homophones. To reduce this problem, n-gram language models are used as they work as spelling correctors. The output logits from the ASR model are fed as input to the language model which then outputs the most probable word based on the previous N-1 words. 

To improve the performance of n-gram language models, we have used modified Kneser-Ney smoothing \cite{smoothing}. Moreover, we have also tuned the hyperparameters (alpha and beta) of the language model to improve performance. We found the optimal value of alpha to be 0.7 and beta to be 0.5. Alpha is the weightage of the language model whereas beta is a penalty term for sequence length. Large alpha values will give more importance to the language model and less importance to the acoustic model. On the other hand, negative beta values will penalize long sequences and make the decoder prefer shorter predictions whereas positive beta values will make the decoder prefer longer candidates.

\subsection{Evaluation Metrics}
\subsubsection{Levenshtein Mean Distance}
The Levenshtein distance \cite{lmd} is a similarity measure between sentences. The distance between two sentences is  There are three methods for editing that can be used. These are:- Insertion, Deletion, and Substitution. The Levenshtein mean distance is simply calculating the total distance for n sentences and then dividing by n.

\subsubsection{Word Error Rate (WER)}
Word error rate is a common and important metric used to measure the performance of ASR models. WER is calculated based on a measurement called Levenshtein distance \cite{lmd}. Levenshtein distance is the minimum number of single-character edits (insertion, deletion, and substitution) required to change one sentence into another. So WER is defined as the Levenshtein distance between the predicted sentence and reference, divided by the number of words in the reference sentence.

\subsubsection{Character Error Rate (CER)}
The character error rate is also calculated based on Levenshtein distance. It is defined as the Levenshtein distance between the predicted sentence and reference divided by the total number of characters in the reference sentence.

\begin{table*}[!ht]
\centering
\caption{Performance of Models}
\label{table:table-1}
\resizebox{\linewidth}{!}{%
\begin{tabular}{|c|c|c|c|c|} 
\hline
\textbf{Model Name}              & \textbf{LM Post Processor}                                                       & \textbf{Transcripts}                                                                                                                                               & \textbf{CER (\%)} & \textbf{WER (\%)}  \\ 
\hline
arijit-pretrained                & 5gram lm model                                                                   & \begin{tabular}[c]{@{}c@{}}30M sentences \\from A\textcolor[rgb]{0.067,0.094,0.153}{I4Bharat }\\\textcolor[rgb]{0.067,0.094,0.153}{IndicCorp dataset}\end{tabular} & 9.88              & 31                 \\ 
\hline
arijit-pretrained-6gram          & \multirow{2}{*}{6gram lm model}                                                  & \multirow{4}{*}{\begin{tabular}[c]{@{}c@{}}100k sentences \\from Common Voice \\Train and Val sets~\end{tabular}}                                                  & 8.16              & 24                 \\ 
\cline{1-1}\cline{4-5}
arijit-finetuned-6gram           &                                                                                  &                                                                                                                                                                    & 1.46              & 4.42               \\ 
\cline{1-2}\cline{4-5}
arijit-finetuned-6gram-tuned     & \begin{tabular}[c]{@{}c@{}}6gram lm model \\(tuned hyperparameters)\end{tabular} &                                                                                                                                                                    & \textbf{1.43}             & 4.24               \\ 
\cline{1-2}\cline{4-5}
arijit-finetuned-15gram-tuned    & \begin{tabular}[c]{@{}c@{}}15gram lm model\\(tuned hyperparameters)\end{tabular} &                                                                                                                                                                    & \textbf{1.43}              & \textbf{4.21}               \\ 
\hline
arijit-finetuned-30M-5gram-tuned & \begin{tabular}[c]{@{}c@{}}5gram lm model\\(tuned hyperparameters)\end{tabular}  & \begin{tabular}[c]{@{}c@{}}100k sentences from\\ Common Voice + 30M\\ sentences from AI4Bharat \\IndicCorp dataset\end{tabular}                                                                         & 1.54              & 4.66               \\
\hline
\end{tabular}
}
\end{table*}

\section{Results}
Table-\ref{table:table-1} shows the performance comparison of various models. "arijit-pretrained" is the current existing SOTA model and it has a WER of 31\% and CER of 9.8\% in the validation set of the common voice dataset. However, if we change the default post-processing language model with a 6-gram language model (trained on the transcripts of common voice dataset), then the WER reduces to 24\% and the CER reduces to 8.16\% which is depicted in the table as "arijit-pretrained-6gram". Next, finetuning the pretrained model on the common voice train set and then using the same 6-gram language model gives WER of 4.42\% which is a 20\% improvement in performance ("arijit-finetuned-6gram"). The CER also reduces to 1.46\% which is a significant decrease from the previous value of 8.16\%. Furthermore, if we finetune the hyperparameters "alpha" and "beta" of the language model, the WER goes down to 4.24\% and the CER becomes 1.43\% ("arijit-finetuned-6gram-tuned"). Apart from that, adding a 15-gram language model as a post-processor to the finetuned arijit model results in a WER of 4.21\% which is a slight improvement while the CER remains the same ("arijit-finetuned-15gram-tuned"). 

Lastly, to create a generalized model, we added a 5-gram language model with the finetuned arijit model labeled as "arijit-finetuned-30M-5gram-tuned" in Table-\ref{table:table-1}. The language model has been trained on the transcripts of common voice and 30M randomly-selected transcripts of the AI4Bharat IndicCorp \cite{indic} dataset. As a result, this model will give the best result on any random dataset not biased towards the common voice dataset. However, the generalized model has a WER of 4.66\% and a CER of 1.54\% on the validation set of the common voice dataset which is not far off the best model we trained.

\section{Discussion}
\subsection{Creating higher order n-gram language models}
We have tried out various order n-gram language models to assess performance gains. All the language models created for this experiment were trained on the transcripts of the common voice train and validations set. We have noticed significant performance gains up to an order of 6 (6-gram). However, after that, the rate of improvement significantly decreases. This might be since there are 7.12 words on average per sentence in the common voice corpus and so 7-gram or higher grams cannot drastically improve the performance. 

\subsection{Large size of n-gram language models}
The size of n-gram language models increases exponentially with the order (gram) of the model we select and so selecting the optimal order is crucial. For the generalized model ("arijit-finetuned-30M-5gram-tuned") in Table-\ref{table:table-1}, we have trained a 5-gram language model on 30.1M sentences. Such a huge transcript generates a language model with a size of 3.5GB for 5-gram. However, the size of the language model increases by more than 20 times if we create a 6-gram language model with the same transcripts with no major improvement in performance. Moreover, the inference time also increases significantly. So, it is pivotal to generate a language model with an optimal order value that balances the performance gain, size, and inference time of the model.

\section{Conclusion}
In this paper, we successfully finetune a SOTA pretrained ASR model on Bangla common voice speech dataset to develop a robust ASR model. To improve performance, we have used an n-gram language model as a post-processor, did hyperparameter tuning on both the language model and the ASR model, and also performed the required preprocessing to ensure a generalized model is generated. In the end, we have a model that has a WER of 4.66\% and a CER of 1.54\% on the validation set of common voice and a Levenshtein mean distance score of 1.65 on the private test set of "DL Sprint - BUET CSE Fest 2022" Kaggle competition.

\section*{Acknowledgment}
The authors would like to thank NSU-Apurba R\&D lab and all affiliated members for their supervision and allocation of resources to train the models.



%

\bibliographystyle{IEEEtran}
\bibliography{bibfile}

\end{document}